\documentclass[runningheads]{llncs}
\usepackage[utf8]{inputenc}

\usepackage[linesnumbered,ruled,vlined]{algorithm2e}
\usepackage{amsmath}
\usepackage{amssymb}
\usepackage{tabularx}

\usepackage{graphicx}

\usepackage{xspace}

\begin{document}

\title{Vision Transformers for Computer Go}

\author{Amani Sagri\inst{1} \and
Tristan Cazenave\inst{1} \and
Jérôme Arjonilla\inst{1} \and
Abdallah Saffidine\inst{2}
}

\institute{
LAMSADE, Université Paris Dauphine - PSL, CNRS, Paris, France
\and The University of New South Wales, Sydney, Australia
}

\maketitle

\begin{abstract}

Motivated by the success of transformers in various fields, such as language understanding and image analysis, this investigation explores their application in the context of the game of Go. In particular, our study focuses on the analysis of the Transformer in Vision. Through a detailed analysis of numerous points such as prediction accuracy, win rates, memory, speed, size, or even learning rate, we have been able to highlight the substantial role that transformers can play in the game of Go. This study was carried out by comparing them to the usual Residual Networks.

\end{abstract}

\section{Introduction}


Due to a huge game tree complexity, the game of Go has been an important source of work in the perfect information setting. In 2007, search algorithms have been able to increase drastically the performance of computer Go programs \cite{Coulom2006,Coulom2007,Kocsis2006,Gelly2011AI}. In 2016, a groundbreaking achievement occurred when AlphaGo became the first program to defeat a skilled professional Go player \cite{Silver2016MasteringTG}. Currently, the level of play of such algorithms is far superior to those of any human player \cite{Silver2016MasteringTG,silver2017mastering,silver2018general}. 

Over the years, various significant advances have been made to improve performance in the game of Go \cite{Cazenave2018Residual,wu2018multilabeled,wu2019strength,wu2019accelerating,wu2020accelerating}. 
Many of these innovations find their roots in other domains, notably in computer vision, where the recognition and interpretation of the Go board's image serve as fundamental inputs. Algorithms such as ResNet \cite{he2016deep,Cazenave2018Residual} and MobileNet \cite{howard2017mobilenets,Cazenave2022Mobile,Cazenave2021Improving} have demonstrated exceptional performance by harnessing groundbreaking developments in computer vision.
However, it is worth noting that one remarkable advancement in the realm of computer vision remains relatively untapped for Computer Go: \emph{transformers}~\cite{10.5555/3295222.3295349}.

Transformers represent a groundbreaking leap in deep learning, reshaping how various tasks in natural language processing (NLP), computer vision, and beyond are approached. Initially developed for NLP tasks, transformers introduce a departure from conventional sequential methods by employing self-attention mechanisms. These mechanisms simultaneously capture intricate interdependencies among all elements in a sequence. This ability to understand nuanced relationships over long distances, without relying on recurrent or convolutional structures, has propelled transformers to the forefront of AI research. Notably, transformers have not only advanced language understanding, exemplified by models like BERT \cite{devlin-etal-2019-bert}, but have also expanded their utility to image analysis, as seen in Vision Transformers (ViTs) \cite{dosovitskiy2020image} and other transformer-based models.
EfficientFormer~\cite{li2022efficientformer}, a transformer-based model, achieves high performance and matches MobileNet's speed on mobile devices, proving that well-designed transformers can deliver low latency in computer vision tasks.

In this paper, we propose to analyze the impact of using Transformer methods in the game of Go. To do this, we use the EfficientFormer architecture. Our study analyses were done in comparison with other state-of-the-art vision architectures in Go such as Residual Networks on a wide range of criteria including prediction accuracy, win rates, memory, speed, architecture size, and even learning rate. Thanks to that, we observe that EfficientFormer is better than Residual Networks on CPU and plays on par on GPU. 

We introduce Computer Go in Section~\ref{sec:notation} and the network architectures used throughout the paper in Section~\ref{sec:algo}.
In Section \ref{sec:experiment}, we present our results and the last section summarizes our work and future work.

\section{Computer Go}
\label{sec:notation}

The game of Go is a turn-taking strategic board game of perfect information, played by two players.
One player adds black stones to a vacant intersection of the board and the opponent adds white stones.
After being placed, a player's stones cannot move.
A group of contiguous stones is removed if and only if the opponent surrounds the group on all orthogonally adjacent points.
The players aim at capturing the most territory and the game ends when no player wishes to move any further.

There exist multiple rules for scoring. We have used the Chinese rule in our experiments: the winner of the game is defined by the number of stones that a player has on the board, plus the number of empty intersections surrounded by that player's stones and komi (bonus added to the second player as compensation for playing second).

Even though the rules are relatively simple, the game of Go is known as an extremely complex one in comparison to other board games such as Chess.
On the standard board of size $19 \times 19$, the number of legal positions has been estimated to be $2.1 \times 10^{170}$.

\begin{figure*}
	\centering
	\includegraphics[scale=0.5]{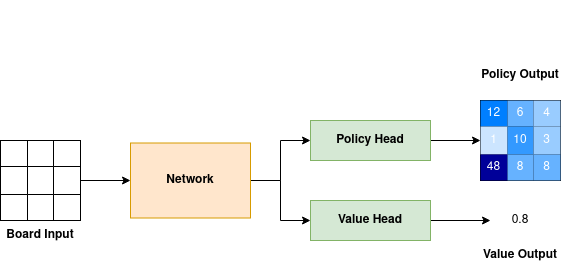}
	\caption{AlphaZero network architecture}
	\label{aznetwork}
\end{figure*}

Algorithms based on Monte Carlo Tree Search (MCTS) \cite{BrownePWLCRTPSC2012} have been achieving excellent performance in the game of Go for many years.
Combining deep reinforcement learning and MCTS as introduced in the \emph{AlphaGo} series programs~\cite{Silver2016MasteringTG,silver2018general,silver2017mastering} has been widely applied. The neural network takes an image of the board as input and produces two outputs: a probability distribution over moves (policy head) and a vector of score prediction for every player (value head) (see Fig.~\ref{aznetwork}). 


\section{Network Architectures}
\label{sec:algo}

\subsection{Residual Network}

Residual Networks are the standard networks for games \cite{Cazenave2018Residual,silver2018general}. They are used in combination with MCTS to evaluate the leaves of the search tree and to give a prior on the possible moves. In order to speed up the computation of the evaluation and of the prior the networks are usually run on a batch of states \cite{cazenave2022batch}.

\begin{figure*}
	\centering
	\includegraphics[width=4cm]{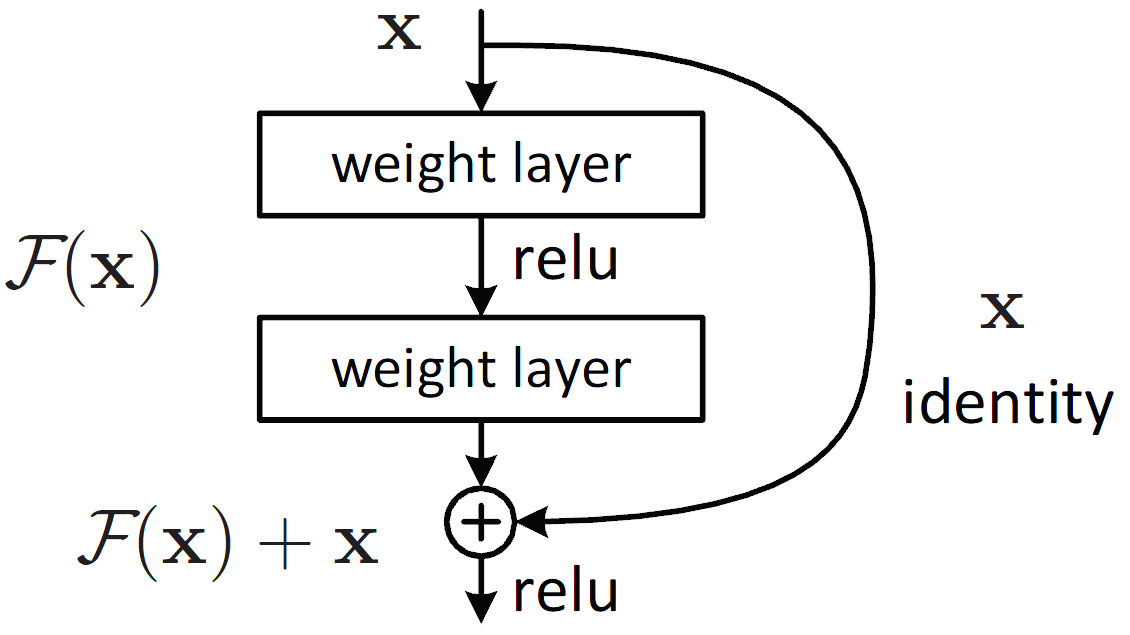}
	\caption{The residual block.}
	\label{residual}
\end{figure*}

The residual layer used for image classification adds the input of the layer to the output of the layer. It uses two convolutional layers before the addition. The ReLU layers are put after the first convolutional layer and after the addition. The residual layer is shown in Figure \ref{residual}. We will experiment
with this kind of residual layer for our Go networks.

\label{sec:net}


\subsection{Transformer}

Transformers are advanced neural network architectures that leverage the concept of self-attention to process and understand complex sequences of data, such as language. Self-attention allows a transformer model to analyze different elements within a sequence and determine their relative importance in relation to one another. By calculating attention scores based on the similarity of these elements, the model can dynamically weigh their significance and understand how they interrelate. Transformers employ multiple self-attention mechanisms (multihead self-attention) operating in parallel, enabling them to capture intricate patterns, dependencies, and contextual nuances across the entire input sequence. 

Transformer was originally proposed as a sequence-to-sequence model \cite{sutskever2014sequence} for machine translation. Later works show that Transformer-based pre-trained models (PTMs) \cite{qiu2020pre} can achieve state-of-the-art performances on various tasks. As a consequence, Transformer has become the go-to architecture in NLP, especially for PTMs. In addition to language related applications, Transformer has also been adopted in CV \cite{parmar2018image,carion2020end,dosovitskiy2020image}, audio processing \cite{dong2018speech,gulati2020conformer,chen2021developing} and even other disciplines, such as chemistry \cite{schwaller2019molecular} and life sciences \cite{rives2021biological}.

If in the field of Natural Language Processing the mechanism of attention of the Transformers tried to capture the relationships between different words of the text to be analyzed, in Computer Vision the Vision Transformers try instead to capture the relationships between different portions of an image.

The mechanism of self-attention, integral to transformers, enables them to excel in tasks ranging from language translation and sentiment analysis to summarization and beyond.

\subsection{Efficient Former}

The EfficientFormer model is a big step forward in making transformer architectures work better for tasks that need real-time results, especially on devices with not much computing power. By adding a dimension-consistent plan, the model can easily switch between different ways of organizing its parts, like in 4D and 3D setups. This way of thinking helps the EfficientFormer model break free from the old rules about how fast transformers can make decisions. This leads to making the time it takes for predictions much shorter. By focusing on making predictions happen fast, a set of EfficientFormer models emerges, each achieving a careful equilibrium between performance and latency. This change in approach reaches its peak with models like EfficientFormer-l1, which impressively demonstrates outstanding top-1 accuracy on benchmarks like ImageNet-1K. At the same time, it manages to keep inference latency remarkably low on mobile devices, aligning closely with the efficiency of optimized versions of MobileNet. The complete range of EfficientFormer models, taken together, significantly underscores the possibilities of tapping into transformers' capabilities for practical real-world uses. 

\begin{figure*}
	\centering
	\includegraphics[width=\textwidth]{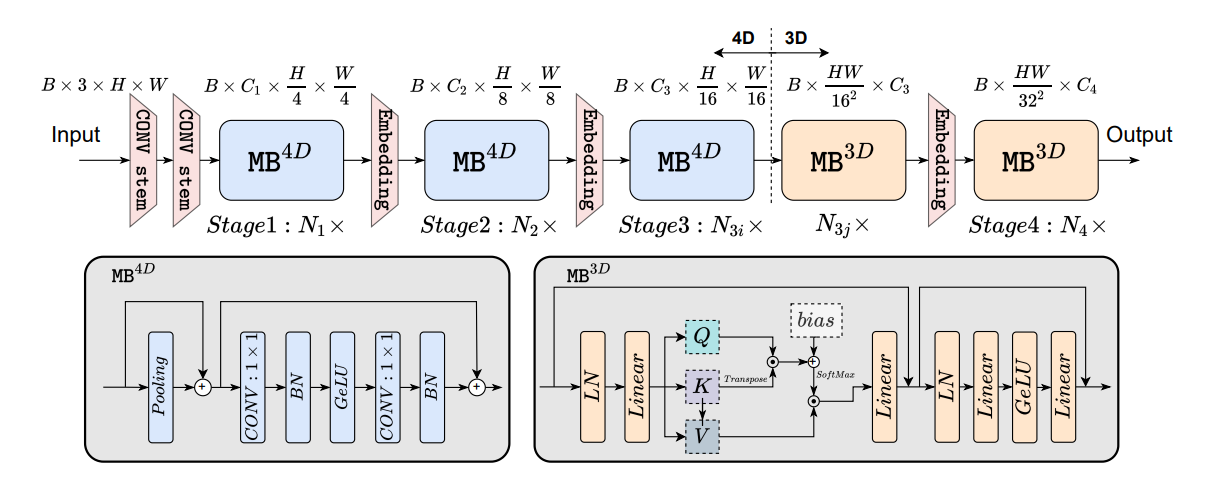}
	\caption{Overview of EfficientFormer architecture \cite{li2022efficientformer}.}
	\label{efficientformer}
\end{figure*}

The network starts with a convolution stem as patch embedding, followed by MetaBlock (MB) as shown by Figure~\ref{efficientformer}. The MB$^{4D}$ and MB$^{3D}$ contain different token mixer configurations, \emph{i.e.}, local pooling or global multi-head self-attention, arranged in a dimension-consistent manner. EfficientFormer is available in various sizes, denoted as l1, l3, l7, and l9. Each size is linked to a tuple of information where the first information is the width and the second information is the depth. The width is a list designing different dimensionalities (number of channels) of the feature vectors processed by different layers and blocks within the neural network. The width represents the number of blocks in different levels of the EfficientFormer architecture. 

The sizes are the following:
\begin{itemize}
    \item  `l1' : ([48, 96],[3, 4])
    \item `l3' : ([64, 128],[4, 6])
    \item `l7' : ([96, 192,],[6, 8])
    \item `l9'; ([128,256],[8,10])
\end{itemize}

\subsection{Adaptation for the game of Go}

In this paper, we took the same work on EfficientFormer model of Li \textit{et al.}~\cite{li2022efficientformer} and adapt the transformer mechanism to the Go game prediction. This necessitated modifying the final layers, which were originally designed for tasks like classification or segmentation, and instead, replacing them with layers tailored for policy and value (\emph{i.e.}, the probability of winning the game) prediction. 

This modification transformed the tasks into a dual output setting, combining multiclass classification and regression functionalities. The value head uses Global Average Pooling followed by two Dense layers \cite{cazenave2019spatial,Cazenave2020Polygames}. The policy head uses a 1x1 convolution to a single plane that defines the convolutional policy \cite{Cazenave2020Polygames}.

Another significant adjustment involved the downsampling and embedding layers commonly used in image classification tasks to detect features by reducing the image size before feeding it into the transformer. However, in the context of Go, the input board's dimensions were fixed at $19\times19$, and it was imperative to preserve this size throughout the training process to avoid losing critical information. Therefore, to retain the richness of the board data, the height and width of the board were maintained during training, ensuring that no valuable details were lost in the process. This tailored architectural approach played a pivotal role in optimizing the models for Go game prediction.

\section{Experimental Results}
\label{sec:experiment}
\subsection{Dataset}

The data used for training comes from the Katago Go program self played games~\cite{wu2019accelerating}. There are 1,000,000 different games in total in the training set. The input data is composed of 31 19x19 planes (color to play, ladders, current state on two planes, two previous states on four planes). The output targets are the policy (a vector of size 361 with 1.0 for the move played, 0.0 for the other moves), and the value (close to 1.0 if White wins, close to 0.0 if Black wins).

\subsection{Experimental Information}

In order to compare the different network architectures we trained them on $500$ epochs. One epoch uses $100,000$ states randomly selected from the Katago dataset with two labels: a one-hot encoding of the Katago move and an evaluation between $0$ and $1$ by Katago of the winrate for White.
The training is done with Adam and cosine annealing \cite{Cazenave2021Cosine} without restarts. Cosine annealing leads to better convergence by modifying the learning rate of Adam. 

In the next tables, we denote Residual(X,Y), the Residual Network of X blocks of Y planes and we denote Efficient(lX), the lX architecture of EfficientFormer. Among the different metrics used, we compute the accuracy, mean squared error (MSE), mean absolute error (MAE), and when possible the winning rate against an opponent. The winning rate is more informative than the other because it combines the impact of improving policy and value network. Accuracy measures the closeness of strategies between the policy network and Katego data.

\subsection{Training and Playing}

Table \ref{tableLR} gives the comparison of some learning rates for small residual and transformer networks. For the Transformer, the best learning rate is observed at $0.002$ whereas for the Residual Network, the best learning rate is observed at $0.0002$. For Residual Networks, the learning rate tends to be significantly lower in comparison to transformers. 

Table \ref{tableGPULatency} gives the latency and peak memory on GPU for the different network architectures we tested. The experiments were carried out on a \emph{RTX 2080 Ti} with $11$ Go of Memory. It is worth mentioning that the Residual Networks and Transformers we examined exhibit similar latency characteristics, however, it can be observed that the memory usage is $3$ times greater with Transformers.

Table \ref{tableCPULatency} gives the number of parameters, the number of evaluations per second, and the CPU latency on an \emph{Epyc} server of the networks we tested. It should be noted that Transformers with far fewer parameters than Residual Networks nevertheless achieve comparable evaluation per second. Furthermore, large Transformers demonstrate superior CPU performance when compared to large Residual Networks.

Table \ref{tableTrain} gives the accuracy, the MSE, and the MAE for the different networks. Additionally, it reveals the winning rate of the l9 Efficient Former when pitted against the competition, assuming either CPU or GPU hardware. For the 256 planes Residual Network the learning rate was set to 0.00005 since the learning rates above could not learn the value. The l9 Efficient Former outperforms its counterparts across various metrics and excels particularly on CPU. When leveraging GPU hardware, it performs at par with the largest Residual Network. 


Lastly, Table \ref{tableBatch} charts the evolution of GPU latency concerning batch size variation for the different network configurations. Large batch sizes are relevant to self-play in Alpha Zero style \cite{silver2018general,wu2019accelerating}. Smaller batch sizes are relevant to normal play with batch parallel MCTS \cite{cazenave2022batch}. This analysis sheds light on how network performance scales with batch size changes. The Residual Networks use relatively more playouts since they parallelize better with current GPU hardware and software.

\begin{table*}[h!]
  \centering
  \caption{Learning rate tuning for different small network architectures over $100$ epochs of $100,000$ states }
  \label{tableLR}
  \begin{tabular}{|lrrrrr|}
\hline
Network & Learning Rate & Batch & Accuracy & MSE & MAE \\
\hline
Residual(10,128) & 0.0008 & 64 & 44.51\% & ~~0.1209 & ~~0.2959 \\
Residual(10,128) & 0.0004 & 64 & 46.25\% & 0.0657 & 0.1927 \\
Residual(10,128) & 0.0002 & 64 & 46.57\% & 0.0642 & 0.1900 \\
Residual(10,128) & 0.0001 & 64 & 45.61\% & 0.0708 & 0.2035 \\
\hline
Efficient(l1) & 0.004 & 64 & 45.72\% & 0.0766 & 0.2097 \\
Efficient(l1) & 0.002 & 64 & 45.89\% & 0.0698 & 0.1973 \\
Efficient(l1) & 0.001 & 64 & 45.43\% & 0.0720 & 0.2022 \\
\hline
  \end{tabular}
\end{table*}

\begin{table*}[h!]
  \centering
  \caption{Latency and peak memory on a RTX 2080 Ti GPU with 11 Go for different architectures and networks of different sizes. The latency and the peak memory are measured using a batch of 64 states. They are averaged over 100 calls to predict after a warmup of 100 previous calls. The latency is the average time in seconds to make a forward pass on a batch of 64 states.}
  \label{tableGPULatency}
  \begin{tabular}{|lrrr|}
\hline
Network & GPU Latency & Evaluations per second on GPU & ~~~~~Peak Memory\\
\hline
Residual(10,128) & 0.0890 & 719 & 436,656,640 \\
Residual(20,128) & 0.0943 & 679 & 350,025,728 \\
Residual(20,256) & 0.1185 & 540 & 452,578,816 \\
Residual(40,256) & 0.1580 & 405 & 529,187,072 \\
\hline
Efficient(l1) & 0.0958 & 668 & 1,101,474,048 \\
Efficient(l3) & 0.1106 & 579 & 1,148,030,976 \\
Efficient(l7) & 0.1307 & 490 & 1,159,418,368 \\
Efficient(l9) & 0.1700 & 376 & 1,179,129,088 \\
\hline
    \end{tabular}
\end{table*}

\begin{table*}[h!]
  \centering
  \caption{Number of parameters and number of evaluations per second on CPU for the different networks.}
  \label{tableCPULatency}
  \begin{tabular}{|lrrr|}
\hline
Network & Parameters & Evaluations per second on CPU & CPU Latency\\
\hline
Residual(10,128)      &  2,967,525 & 23.07 & 0.043 \\
Residual(20,128)      &  5,924,325 & 12.24 & 0.082 \\
Residual(20,256)      & 23,645,029 &  3.29 & 0.304 \\
\hline
Efficient(l1)  &    674,885 & 15.27 & 0.065 \\
Efficient(l3) &  1,381,541 & 13.52 & 0.074 \\
Efficient(l7) &  3,581,573 & 10.90 & 0.092 \\
\hline
    \end{tabular}
\end{table*}

\begin{table*}[h!]
  \centering
  \caption{Comparison of networks for 500 epochs of 100,000 states per epoch and a batch size of 64. The winrate WinCPU is the result of 800 randomized matches on CPU against Efficient(l9) with 10 seconds of CPU per move for both sides. The GPU winrate is calculated by using the same GPU time for both networks. The Accuracy, MSE and MAE were computed on a set of 50,000 states sampled from 50,000 games that were never seen during training.}
  \label{tableTrain}
  \begin{tabular}{|lrrrrrrr|}
\hline
Network & Learning Rate & Batch & Accuracy & MSE & MAE & WinCPU & WinGPU\\
\hline
Residual(10,128) & 0.0002 & 64 & 49.12\% & ~~0.0534 & ~~0.1649 & 33.5\% & 20.4\%\\
Residual(20,128) & 0.0002 & 64 & 50.29\% & 0.0516 & 0.1618 & 31.6\% & 25.8\%\\
Residual(20,256) & 0.00005 & 64 & 52.50\% & 0.0476 & 0.1518 & 30.6\% & 51.0\%\\
Residual(40,256) & 0.00005 & 32 & 51.27\% & 0.0499 & 0.1586 & 8.9\% & 34.7\%\\
\hline
Efficient(l1)  & 0.002 & 64 & 49.35\% & 0.0553 & 0.1659 & 11.6\% & 8.1\%\\
Efficient(l3) & 0.002 & 64 & 51.28\% & 0.0484 & 0.1519 & 31.0\% & 19.4\%\\
Efficient(l7) & 0.002 & 64 & 53.01\% & 0.0440 & 0.1422 & 50.4\% & 38.3\%\\
Efficient(l9) & 0.001 & 64 & 54.29\% & 0.0405 & 0.1351 & - & - \\
\hline
  \end{tabular}
\end{table*}

\begin{table*}[h!]
  \centering
  \caption{Evolution of the A6000 GPU latency with the size of the batch. The latency and the peak memory are the median values of 7 runs. Each run is the average over 100 forwards after a warmup of 100 forwards.}
  \label{tableBatch}
  \begin{tabular}{|lrrrr|}
\hline
Network & Batch & GPU Latency & Evaluations per second on GPU & Peak Memory \\
\hline
Efficient(l9) & 32 & 0.128 & 250 & 589,454,592 \\
Efficient(l9) & 64 & 0.168 & 381 & 1,141,404,672 \\
Efficient(l9) & 128 & 0.224 & 571 & 2,297,159,168 \\
Efficient(l9) & 256 & 0.346 & 740 & 4,359,236,608 \\
Efficient(l9) & 512 & 0.583 & 878 & 8,672,660,992 \\
Efficient(l9) & 1024 & 1.062 & 964 & 17,121,701,376 \\
\hline
Residual(20,256) & 32 & 0.111 & 288 & 253,801,472 \\
Residual(20,256) & 64 & 0.126 & 508 & 548,938,240 \\
Residual(20,256) & 128 & 0.159 & 805 & 800,936,192 \\
Residual(20,256) & 256 & 0.227 & 1128 & 1,566,134,528 \\
Residual(20,256) & 512 & 0.368 & 1391 & 2,954,716,416 \\
Residual(20,256) & 1024 & 0.667 & 1535 & 4,793,448,960 \\
\hline
  \end{tabular}
\end{table*}

\section{Conclusion}

EfficientFormer's architecture showcases remarkable parameter efficiency, especially when compared to the Residual Network architecture, particularly in larger networks. This translates into superior performance on CPU, making it the preferred choice in this domain. Interestingly, when it comes to GPU utilization, both architectures perform at a similar level, especially for the largest networks in our experimentation.

Moreover, it is worth highlighting that the EfficientFormer architecture we explored for Go is not limited to this particular game; it exhibits versatility and applicability to a wide range of other games and domains.



\bibliographystyle{splncs04}
\bibliography{main}

\end{document}